\pdfoutput=1

\documentclass[11pt]{article}

\usepackage{acl}

\usepackage{times}
\usepackage{latexsym}

\usepackage[T1]{fontenc}

\usepackage[utf8]{inputenc}

\usepackage{microtype}

\usepackage{booktabs}
\usepackage{graphicx}
\usepackage{amsmath}
\usepackage{amssymb}
\usepackage{enumitem}
\usepackage{multirow}
\usepackage[export]{adjustbox}

\newlist{todolist}{itemize}{2}
\setlist[todolist]{label=$\square$}
\usepackage{pifont}
\newcommand{\mysubsection}[1]{\subsection{#1}}
\usepackage{soul}
\usepackage{xcolor}
\definecolor{fig1yellow}{rgb}{1.0, 0.9960784314, 0.7607843137}
\definecolor{fig1green}{rgb}{0.7568627451, 0.9960784314, 0.7568627451}
\definecolor{fig1red}{rgb}{0.9960784314, 0.7529411765, 0.7529411765}
\definecolor{fig1blue}{rgb}{0.7568627451, 1.0, 0.9960784314}
\newcommand{\hlc}[2][yellow]{{%
    \colorlet{foo}{#1}%
    \sethlcolor{foo}\hl{#2}}%
}

\title{Asking the Right Questions in Low Resource Template Extraction}

\author{Nils Holzenberger\thanks{~~Equal contributions} \and Yunmo Chen\textsuperscript{$*$} \and Benjamin Van Durme \\
\texttt{\{nilsh,yunmo,vandurme\}@jhu.edu} \\
  Center for Language and Speech Processing \\
  Johns Hopkins University \\
  Baltimore, Maryland, USA}

\begin{document}

\maketitle

\begin{abstract}

Information Extraction (IE) researchers are mapping tasks to Question Answering (QA) in order to leverage existing large QA resources, and thereby improve data efficiency. 
Especially in \textit{template extraction} (TE), mapping an ontology to a set of questions can be more time-efficient than collecting labeled examples.
We ask whether end users of TE systems can design these questions, and whether it is beneficial to involve an NLP practitioner in the process.
We compare questions to other ways of phrasing natural language prompts for TE. 
We propose a novel model to perform TE with prompts, and find it benefits from questions over other styles of prompts, and that they do not require an NLP background to author.

\end{abstract}

\section{Introduction}
\label{sec:introduction}
\begin{figure*}[ht!]
\centering
\includegraphics[width=.95\textwidth]{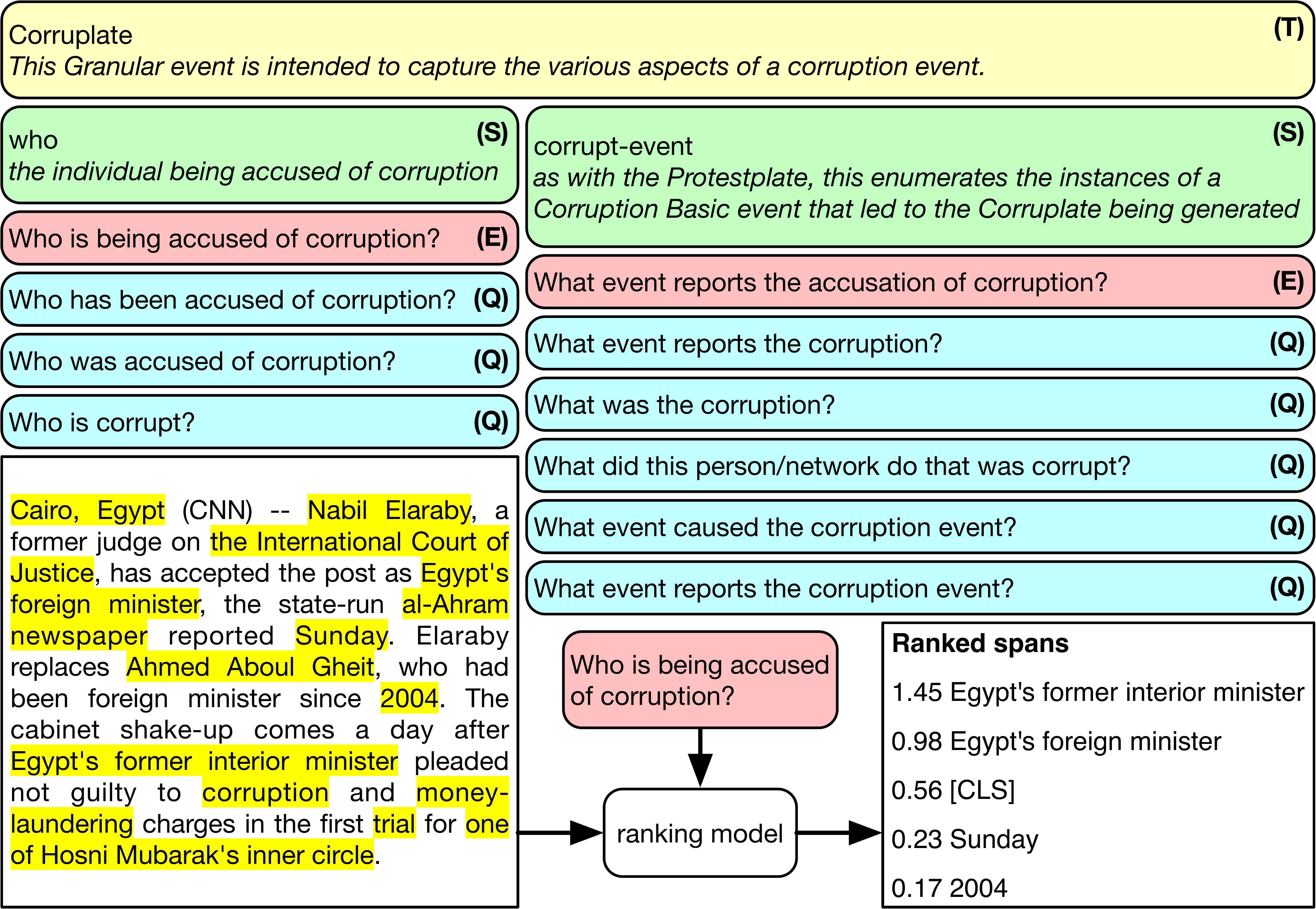}
\caption{Top: Example template from the Granular ontology. Bottom: an example document from the Granular dataset, with spans highlighted. Expected answers for each slot are specified using superscripts. \hlc[fig1yellow]{ (T): template name and description}, \hlc[fig1green]{(S): slot name and description}, \hlc[fig1red]{(E): expert question}, \hlc[fig1blue]{(Q): non-expert question}.}
\label{fig:data}
\vspace{-3mm}
\end{figure*}

Information Extraction (IE) tasks such as event extraction (EE) and template extraction (TE) \citep[\textit{inter alia}]{du21template} are commonly presented as slot filling for template-like structures, with fillers extracted from a textual passage.
Current extraction systems powered by neural networks have achieved state-of-the-art performance but require significant amounts of annotated data for training. 
Given the complexity and specialized domains of many ontologies, annotation processes are often performed by domain experts, making high-quality annotations hard to obtain.

One way to mitigate data scarcity is to use natural language \emph{prompts}.
In the general sense, prompts are natural language (NL) strings which, fed to an IE model, serve as a flexible inductive bias across many tasks \citep{raffel20exploring}.
Prompts have been widely adopted in many NLP tasks~\citep[\textit{inter alia}]{levy17zero,McCannKXS18}, paving the way for formulating TE tasks as Question Answering (QA)~\citep[\textit{inter alia}]{DuC2020,chen20reading}. 

Previous work has focused on different ways to compose prompts~\citep{ShinRLWS20} and approaches to tuning prompts towards better model performance~\citep{logan21cutting}.
However, to the best of our knowledge, under the TE as QA (TE-QA) formulation, a thorough study has not been pursued to investigate the effectiveness of prompts formulated as NL questions, and humans' ability to judge the usefulness of a given prompt.
In particular, we investigate whether potential end users of a TE system can be allowed to drive task definitions. 
For that, we let humans who have no background in TE write prompts in the form of NL questions, as an efficient way of annotating for the task.
Since end users are not experts in TE modeling, and are likely more used to writing queries used in search engines, we are concerned with whether a QA-based TE system can perform well with their questions~---~a realistic setting for people using such IE systems.
With the advancement of QA models, we would like to investigate the robustness of such systems under different types of prompts.
Furthermore, we quantify whether end users are able to intuitively distinguish between higher- and lower-quality prompts.\footnote{~We define higher quality prompts as prompts yielding higher model performance.}
The contributions of this work are as follows:
\begin{enumerate}[leftmargin=*,topsep=0pt,itemsep=0pt,partopsep=0pt,parsep=0pt]
    \item We design an analysis framework to empirically evaluate the performance of TE-QA systems fed with different prompts: we propose a ranking-based TE-QA model that bypasses the need to extract spans, to focus on examining the effectiveness of various prompts.
    \item We collect prompts from potential end users, and thoroughly analyse prompt quality.
    \item We conclude that: a) the phrasing of prompts \textit{matters}; b) prompt writers without TE modeling background \textit{can} write prompts leading to high model performance; c) humans are \textit{unable} to gauge the quality of a prompt.
\end{enumerate}

\section{Background}
\label{sec:related-work}
\paragraph{TE as QA} 

The terms QA and MRC have both been used to describe the use of a linguistic description to query a context in order to extract answers.
Under its formulation, prompts\footnote{~In this case, prior work has been using prompt, query, and question interchangeably.} serve as the key component that elicits information and guides the extraction process.
QA has been widely explored in different NLP tasks~\citep{McCannKXS18}, such as semantic role labeling~\citep{ZettlemoyerFHM18}, dialogue state tracking~\citep{GaoSACH19}, and fact verification~\citep{WebnerMMLR19}.

Recent work adopts the formulation for various TE tasks, and achieves state-of-the-art performance~\citep[\textit{inter alia}]{LiYSLYZL19,WuWYWL20,LiFMHWL20,DuC2020}.
Interestingly, those works make use of different styles of prompts: hand-written NL questions~\citep{LiYSLYZL19}, a series of NL questions instantiated from a set of templates~\citep{DuC2020}, descriptions from annotation guidelines rewritten as bleached definitions~\citep{chen20reading}, and mentions detected using a Conditional Random Field along with surrounding context used as questions~\citep{WuWYWL20}.
These forms of prompts work and, to some extent, serve their modeling purposes well. However, there is a lack of analysis on how much the form and the phrasing of the prompt matters for model performance. 
Further, it is not clear whether end users can judge the quality of a prompt in terms of expected model performance. 

\paragraph{Prompt engineering}

NL prompts contain information not captured by model parameters, hence serving as a flexible way to inject inductive bias, and playing the role of a hyper-parameter~\citep{GPT3paper}.
Unlike other hyper-parameters, given the huge discrete search space, searching for an optimal NL prompt that maximizes model performance is an NP-hard problem.
Prior work~\cite[\textit{inter alia}]{ShinRLWS20, schick21exploiting, schick21it} investigates searching for improved NL prompts with respect to model performance using gradients as signals. 
These approaches usually lead to non-fluent NL prompts, e.g. a list of keywords. 
We are concerned with understanding how robust a model is to prompts written by humans, and whether performance benefits from a prompt being authored by an NLP expert.
We rely on human annotators to explore the space of NL prompts.

Another line of work~\citep[\textit{inter alia}]{qin21learning,LiuZDDQYT21} abandons NL questions and optimizes soft prompts in a continuous space.
It allows training the prompts along with model parameters and bypasses the difficult search problem.
Although soft prompts may perform similarly or better than explicit prompts, the tuning process is no longer interpretable for humans. Moreover, soft prompts themselves are no longer interpretable for humans. In low resource scenarios, where inductive biases play an important role in compensating data scarcity, NL prompts can better capture human-provided inductive biases. 

\paragraph{Low resource}

\citet{levy17zero,chen20reading}~(\textit{inter alia}) conduct experiments in zero- or few-shot learning.
They provide insights on  task performance and generalization ability under such settings, but without attention to data efficiency.
\citet{lescao21how} quantify the effect of providing a prompt as opposed to more training data. \citet{logan21cutting} contrast tuning the prompt and tuning the model itself.

\paragraph{Templates and events}

We define a \emph{template} as a semantic structure that consists of \emph{slots} and represents an event or a conjunction of events. 
The definition shares similarities with that of \emph{events} used throughout IE, where \emph{slot} is a counterpart to \emph{argument}, allowing us to recast existing IE datasets to the framework of templates and slots.

In prior work, event semantics has been investigated in different settings.
ACE is a \emph{sentence-level} event task~\citep{ACE2005} which requires the detection of event trigger and arguments residing within the same sentence.
RAMS~\citep{EbnerXCRD20} contains events with arguments spanning across \emph{multiple sentences}.
The Granular task proposed by the BETTER program\footnote{~\href{https://www.iarpa.gov/research-programs/better}{https://www.iarpa.gov/research-programs/better}} further extends the paradigm to combine \emph{multiple simple sentence-level events, entities, factuality, and their temporal relationships} to form storylines (named ``templates''). 
Although these datasets differ in complexity, they can fit under the same template framework.

We are interested in how prompt phrasing affects a model's NL understanding ability in both high and low resource scenarios. 
Given the above definition of \emph{templates}, we use the task of filling slots in templates~\citep[\textit{inter alia}]{du20document,chen20reading,du21grit,du21template} to examine the model, allowing us to evaluate template understanding with ontologies of different complexities.
For the aforementioned event datasets, we show statistics in Table~\ref{tbl:datasets}.

\begin{table}[h]
  \centering
  \adjustbox{max width=\linewidth}{
  \begin{tabular}{llccc}
  \toprule
  & Split & \bf ACE 2005 & \bf RAMS & \bf Granular\\
  \midrule
  \#Template types & & 33 & 139 & 4 \\
  \#Slot types & & 22 & 65 & 63\textsuperscript{*} \\
  \multirow{3}{*}{\#Templates/\#Slots} & train & 4202/4859 & 7329/17026 & 252/1449 \\
  & dev & 450/605 & 924/2188 & 56/294 \\
  & test & 403/576 & 871/2023 & 47/269 \\
  \bottomrule
  \end{tabular}
  }
  \caption{Dataset statistics. \textsuperscript{*} indicates that we only consider slots that have spans as fillers.}
  \label{tbl:datasets}
\end{table}

\section{Prompt collection}
\label{sec:writing-questions}
In the ontologies associated with Granular, ACE and RAMS, each template comes with an NL description.  These descriptions are meant to help humans understand the scope of the ontology, and to help when annotating documents by hand.
Slots may or may not have a linguistic description.
For all three datasets, we asked annotators to write questions that captured the information stated in the ontology description.

\mysubsection{Annotation protocol} 
First, one expert NLP practitioner wrote a question for each slot, referred to as ``expert''.\footnote{~For our purposes, experts are people who know how QA models work, and have a sense of how to inject detailed information via prompts.}
Second, we recruited non-expert annotators to write a question for each slot in a given template.\footnote{~Annotators were paid undergraduate students without background in the task nor in NLP modeling.} These annotators have no background in TE, and we view them as potential end users of the TE system.
We provided the annotators with the name of the template in the ontology\footnote{~e.g. ArtifactExistence:ArtifactFailure:MechanicalFailure}, the description of the template, and if it exists, the description of the slot.
Examples are shown in Figure~\ref{fig:data}.
Annotators were also shown some templates with the expert questions already filled in, to help them get started.
The templates, as well as the questions within a template, were presented in a random order, mitigating any variations in the quality of the annotations due to seeing one slot at an earlier or later point in the annotation process.
For the Granular ontology, 8 annotators were hired. Each produced questions for 3 templates. Templates were assigned to annotators so that all 4 templates in the ontology were annotated equally, resulting in 6 questions for each slot. Some questions were identical across multiple annotators, resulting in less than 6 distinct questions for some slots, as can be seen in Figure~\ref{fig:data}.
Similarly, 5 annotators worked on the ACE and RAMS ontologies, resulting in 3 and 1 questions per slot, respectively.

\mysubsection{Prompt aggregation}

With $n$ questions per slot, we group questions across slots into $n$ series (shown as \textsc{Series} in the following tables).
Each series covers the entirety of the dataset's templates and contains questions that come from the smallest number of different annotators possible.
This is done by assigning an index to each annotator. For each slot, questions are ordered by increasing index. The $i$-th question is assigned to the $i$-th series.
Variation in performance across series is thus kept as close as possible to variations in performance across annotators.
Examples are shown in Figure~\ref{fig:data}, and Tables~\ref{tbl:question-samples} and ~\ref{tbl:scoring}.

To measure the diversity of the questions collected, we compute pairwise similarities of the questions filling the same slot using SBERT~\citep{reimers19sentence}.
We report detailed statistics in Table~\ref{tbl:question-statistics-1} and show examples in Table~\ref{tbl:question-statistics-2}.
Pairwise similarity of questions is generally high.
Including expert questions does not change the similarities much, indicating a certain amount of information being shared in the embedding space.
RAMS questions show lowest absolute similarity.
We hypothesize that more fine-grained ontologies such as used in RAMS make annotators write diverse questions to distinguish subtle differences. For example, the AIDA ontology used for RAMS distinguishes between one-way and two-way communication, and whether communication happens in person or not.\footnote{~e.g. Contact:CommandOrder:Broadcast, Contact:CommandOrder:Correspondence and Contact:CommandOrder:Meet}

Manual analysis of questions reveals that annotators tended to write questions that allow for a narrower interpretation than the expert's, as can be seen in Table~\ref{tbl:question-samples}.
We attribute this to the expert's effort to formulate questions close to the meaning of the ontology, but open enough to accommodate a wide range of answers, in an attempt to balance precision and recall.
Table~\ref{tbl:question-statistics-3} shows the statistics of questions written by experts and non-experts.
We consider ACE and Granular simpler than the AIDA ontology, as they have fewer template types and slot types (shown in Table~\ref{tbl:datasets}).
For these simpler ontologies, the expert wrote shorter, somewhat less specific questions.
For the complex AIDA ontology, the expert wrote longer questions, to highlight subtle differences between templates. This was sometimes done at the expense of fluency.

\begin{table}[h]
\small
\centering
\begin{tabular}{l|l|l}
\toprule
 & with expert & without expert \\
\midrule
ACE & $89.1 \pm 6.5$ & $88.8 \pm 6.6$ \\
Granular & $90.5 \pm 4.4$ & $90.2 \pm 4.4$ \\
RAMS & $87.4 \pm 8.8$ &  \\
\bottomrule
\end{tabular}
\caption{Pairwise cosine similarity (in \%) of questions written for the same slot, formatted as \emph{average} $\pm$~\emph{standard deviation}. ``With expert'' includes the expert question in the pool of questions.}
\label{tbl:question-statistics-1}
\vspace{-5mm}
\end{table}

\newcommand{\myindent}[0]{\hspace{0.05\linewidth}}
\begin{table}[h]
\small
\centering
\begin{tabular}{p{\linewidth}}
\toprule
\textbf{ACE, similarity=0.7:} \\
\myindent For what was the defendent being tried? \\
\myindent \emph{What criminal charges is the trial for?} \\
\midrule
\textbf{Granular, similarity=0.8:} \\
\myindent How much was this person fined? \\
\myindent What monetary punishment is the individual facing? \\
\midrule
\textbf{RAMS, similarity=0.9:} \\
\myindent \emph{Where did the money transfer take place?} \\
\myindent Where did the transaction occur? \\
\bottomrule
\end{tabular}
\caption{Example questions and their similarities from different datasets. Expert questions are in \emph{italics}.}\label{tbl:question-statistics-2}
\vspace{-5mm}
\end{table}

\begin{table}[h]
\small
\centering
\begin{tabular}{l|ll|ll}
\toprule
            & \multicolumn{2}{l}{non-expert} & \multicolumn{2}{l}{expert} \\
\midrule
ACE  & $5.3 \pm 1.4$ & (6) & $5.1 \pm 1.5$ & (6) \\
Granular    & $7.1 \pm 2.2$ & (5) & $6.5 \pm 2.0$ & (7) \\
RAMS        & $5.7 \pm 1.8$ & (4) & $7.0 \pm 2.4$ & (7) \\
\bottomrule
\end{tabular}
\caption{Question length in words, formatted as \emph{average}~$\pm$~\emph{standard deviation} (\emph{median}).}\label{tbl:question-statistics-3}
\vspace{-5mm}
\end{table}

\begin{table}[h]
\small
\centering
\setlength{\tabcolsep}{0pt}
\begin{tabular}{p{\linewidth}}
\toprule
\textsc{Life:Injure}.]\ An \textsc{Injure} Event occurs whenever a \textsc{Person} entity experiences physical harm. \textsc{Injure} Events can be accidental, intentional or self-inflicted. \\
\textsc{Instrument-Arg}. The device used to inflict the harm. \\
\multicolumn{1}{c}{\rule{0.5\linewidth}{0.05 pt}}\\
\emph{What weapon or vehicle was used to injure?} \\
What device was used to cause the injury?  \\
What weapon was used to harm the person? \\
What device was used to inflict harm? \\
\midrule
\resizebox{\linewidth}{!}{\textsc{Movement:TransportArtifact:SmuggleExtract}.}
The smuggling or extracting of an artifact by an agent out of one place into another.
<Transporter> transported <Artifact> in <Vehicle> from <Origin> place to <Destination>~place. \\
\textsc{Transporter}. \\
\multicolumn{1}{c}{\rule{0.5\linewidth}{0.05 pt}}\\
\emph{Who transported the artifact that was smuggled or~extracted?}  \\
Who smuggled or extracted an artifact? \\
\bottomrule
\end{tabular}
\caption{Example questions written for the ACE (top) and RAMS (bottom) ontologies. The expert question is in \emph{italics}.}\label{tbl:question-samples}
\vspace{-3mm}
\end{table}

\section{Model}
\label{sec:model}
To answer the aforementioned questions, we would like our QA model performance to depend, as much as possible, only on prompts used, excluding the effect of other factors such as finding candidate spans for slot filling. We propose a learning-to-rank model that consumes a given set of candidate spans and a prompt, and picks the most probable slot fillers.

\subsection{Problem formulation}

Let $C$ be a text passage, or context.
We refer to a specific template and slot type as $y$.
The set of candidate spans $A$ in $C$ are those to be considered when predicting $U_y$, the subset of $A$ that fills slot $y$.
We also assume access to a span in $C$ that triggers the template.\footnote{A trigger span in IE is the key textual evidence that supports prediction of an event or template.}
The candidate spans $A$ are given as part of the TE datasets.

We use a \emph{prompt} to query the model to predict a set of spans to fill the slot $y$.
We denote the prompt as ${Q_y=(q_1, \cdots, q_m)}$, and the text passage as ${C=(c_1, \cdots, c_n)}$.
The model predicts a set of spans $\overline{U}_y$ where $\overline{U}_y \subset A$. 
Our approach is illustrated in Figure~\ref{fig:data}.

\subsection{Context and span encoding}
\label{subsec:architecture}

We adopt an encoder-based machine reading architecture to jointly encode a prompt $Q_y$ and passage $C$~\citep{DevlinCLT19}.
The prompt $Q_y$ and the text $C$ are concatenated with special delimiters\footnote{~We use BERT delimiters in this description, but the actual delimiters might differ depending on the encoder design.} and passed through the underlying encoder:
\begin{equation}
    \resizebox{\linewidth}{!}{$
    \textsc{Encoder}(\left [ \textsc{cls}, q_1, \cdots, q_m, \textsc{sep}, c_1, \cdots, c_n, \textsc{sep} \right ]) \nonumber
    $}
\end{equation}
where $\textsc{cls}$ is a special token whose embedding represents the whole sequence, and $\textsc{sep}$ serves as the sequence separator.
We denote the output encoding of each question token ${q_i ~(1 \le i \le m)}$ as ${\mathbf{q}_i \in \mathbb{R}^d}$ and the encoding of each text token ${c_i ~(1 \le i \le n)}$ as ${\mathbf{c}_i \in \mathbb{R}^d}$.
We concatenate a positional embedding ${\mathbf{p}_i \in \mathbb{R}^p ~(1 \le i \le n)}$ to each $\mathbf{c}_i$, marking the distance in tokens of $c_i$ to the tokens representing the template trigger.
The distances are non-negative integers, so that the ${\mathbf{p}_i}$ are learned embeddings.
We consider the set of candidate spans ${A = \left \{a_1, \cdots, a_k \right \}}$,  where ${a_k=(a_{k}^{\textrm{start}}, a_{k}^{\textrm{end}})}$ and ${1 \le a_{k}^{\textrm{start}} \le a_{k}^{\textrm{end}} \le n}$.

\subsection{Span ranking and dynamic threshold}

As slots can have any number of filler spans, only picking the most probable span~\citep{DevlinCLT19} cannot suffice.
Besides, some slots might not have any fillers.
Prior work such as \citet{DevlinCLT19} uses the \textsc{cls} token to determine whether the question is answerable. 
We adopt the use of \textsc{cls}, and extend it with the idea of dynamic threshold from \citet{ChenCD20}, to be able to support any number of answer spans.
Similar to \citet{LeeHLZ17}, we compute a score $s(a_k)$ for each candidate span $a_k$, indicating how likely it belongs to $U_y$:
\begin{equation}
     s(a_k)=\textsc{ffnn}([c_{a_{k}^{\textrm{start}}}, c_{a_{k}^{\textrm{end}}}]) \nonumber
\end{equation}
where $[c_{a_{k}^{\textrm{start}}}, c_{a_{k}^{\textrm{end}}}]$ is the concatenation of the two vectors representing the start and end tokens of $a_k$, and \textsc{ffnn} is a feed-forward neural network.
We score the $\textsc{cls}$ token the same way as other candidate spans to obtain $s(\textsc{cls})$.
It then serves as a sentinel to dynamically separate relevant spans from irrelevant spans.
The underlying ranking relation to be learned is:
\begin{equation}
    u \succ \textsc{CLS} \succ v, \quad \forall u \in U_y, \forall v \in V_y = A \setminus U_y \nonumber
\end{equation}
where $\succ$ means ``ranks higher than''.
For model training, we employ a hinge ranking loss:
\begin{alignat}{10}
    J_{u \succ \textsc{cls}} & = & \displaystyle\frac{1}{|U_y|}\sum_{u \in U_y} & [ & \alpha & - & ( s(u) & -  & s(\textsc{cls}) )   & ]_+ \nonumber \\
    J_{\textsc{cls} \succ v} & = & \displaystyle\frac{1}{|V_y|}\sum_{v \in V_y} & [ & \beta & + & ( s(v) & -  & s(\textsc{cls}) ) & ]_+ \nonumber
\end{alignat}
\begin{equation}
J = \lambda J_{u \succ \textsc{cls}} + (1-\lambda) J_{\textsc{cls} \succ v} \nonumber
\end{equation}
where the hyper-parameter $\lambda$ controls the trade-off between both losses, and hyper-parameters $\alpha$ and $\beta$ are the margins of the loss functions.
During inference, we first compute scores for all candidate spans and $\textsc{cls}$.
The predicted answer set is then decoded as ${\overline{U}_y = \left \{ a \mid s(a) \geq s(\textsc{cls}), \forall a \in A \right \}}$.

\section{Experiments}
\label{sec:experiments}
\mysubsection{Experimental settings}
\label{subsec:setting}

We run experiments on the Granular, ACE and RAMS datasets to measure the impact of using different prompts on TE-QA.
We use Unified-QA~\citep{khashabi20unifiedqa} as the \textsc{Encoder} described in Section~\ref{subsec:architecture}.
Since our experiment setting is inherently the same as by \citet{EbnerXCRD20} and \citet{ChenCVD20}, we follow prior work and use span-level F1 as the evaluation metric. 
We take \citet{ChenCVD20} as our baseline, shown as \textsc{JointArg} in the tables, since it obtains previous state-of-the-art results on both ACE and RAMS using gold mention spans.
As far as we are aware, we are the first to publish results on the Granular dataset with this experimental setup.

For each set of prompts, we use Optuna~\citep{akiba19optuna} to tune hyper-parameters, running 20 experiments per set.
Each experiment consists of 3 training runs with 3 different random seeds.\footnote{~Except for RAMS, where we use single runs.}
During the hyper-parameter sweep, we pick the best performing model on the validation set based on micro-averaged F1 scores.
With the best set of hyper-parameters, we retrain with 5 distinct random seeds and report the test scores by taking the average of 5 runs.
The process intends to minimize the effect of hyper-parameters other than the choice of prompts, hence illustrating the best possible performance for each set of prompts.
Therefore, we are able to make a fair comparison across different types of prompts.
Further details can be found in Table~\ref{tbl:hyperparameters} in the Appendix.

\mysubsection{Results and discussion}
\label{subsec:results}

We report results in Tables~\ref{tbl:granular}, \ref{tbl:ace} and \ref{tbl:rams}.
We compare expert questions (\textsc{Expert}), questions collected in Section~\ref{sec:writing-questions} (\textsc{Series}), and the following different types of prompts:
\begin{itemize}[leftmargin=*,topsep=0pt,itemsep=0pt,partopsep=0pt,parsep=0pt]
    \item \textsc{SpecialTokens}, a single token per slot, providing a randomly initialized, trainable embedding, seen as a naive soft prompt baseline in the spirit of \citet{qin21learning},
    \item the \textsc{Name} of the template and slot in the ontology, separated by a space,
    \item a fluent \textsc{Description} of the template and slot, when available in the ontology.
\end{itemize}

\begin{table}[h]
\small
\centering
\begin{tabular}{l|c|c}
\toprule
Prompt & Micro-F1 & Macro-F1 \\
\midrule
\textsc{SpecialTokens} & $35.3 \pm 3.0$ & $35.5 \pm 2.8$ \\
\textsc{Name} & $55.8 \pm 0.4$ & $\mathbf{57.3} \pm 1.0$ \\
\textsc{Description} & $60.6 \pm 0.7$ & $53.0 \pm 1.6$ \\
\textsc{Expert} & $\mathbf{61.5} \pm 0.3$ & $55.6 \pm 1.1$ \\
\textsc{Series-1} & $57.9 \pm 0.7$ & $54.8 \pm 1.3$ \\
\textsc{Series-2} & $\mathbf{61.2} \pm 0.5$ & $56.6 \pm 0.9$ \\
\textsc{Series-3} & $57.5 \pm 0.5$ & $49.9 \pm 1.1$ \\
\textsc{Series-4} & $61.0 \pm 0.4$ & $\mathbf{59.5} \pm 1.2$ \\
\textul{\textsc{Series-5}} & $\mathbf{62.1} \pm 0.4$ & $\mathbf{57.9} \pm 0.9$ \\
~~- \textsc{Pretraining} & $\mathbf{63.8} \pm 0.3$ & $\mathbf{58.1} \pm 1.2$ \\
\textsc{Series-6} & $\mathbf{61.6} \pm 0.5$ & $56.3 \pm 0.9$ \\
\textsc{Best} & $59.6 \pm 0.7$ & $\mathbf{58.0} \pm 1.0$ \\
\textsc{Worst} & $\mathbf{61.2} \pm 1.0$ & $55.8 \pm 1.4$ \\
\bottomrule
\end{tabular}
\caption{Test scores on Granular dataset with standard error. Best scores and scores that would be the best by changing 1 standard error are bolded. 
The question set with the highest dev micro F1 is \textul{underlined}.}\label{tbl:granular}
\end{table}

\begin{table}[h]
\small
\centering
\begin{tabular}{l|l|l}
\toprule
Prompt & Micro-F1 & Macro-F1 \\
\midrule
\textsc{JointArg\textsuperscript{*}} & $65.6$ & - \\
\textsc{SpecialTokens} & $65.1 \pm 1.5$ & $72.6 \pm 0.6$ \\
\textsc{Name} & $\mathbf{82.0} \pm 0.3$ & $\mathbf{86.3} \pm 0.2$ \\
\textsc{Description} & $80.8 \pm 0.2$ & $85.9 \pm 0.2$ \\
\textsc{Expert} & $\mathbf{82.1} \pm 0.2$ & $\mathbf{86.2} \pm 0.2$ \\
\textsc{Series-1} & $\mathbf{82.0} \pm 0.4$ & $\mathbf{86.4} \pm 0.5$ \\
\textul{\textsc{Series-2}} & $\mathbf{82.3} \pm 0.3$ & $\mathbf{86.7} \pm 0.4$ \\
~~- \textsc{Pretraining} & $51.6 \pm 2.9$ & $68.8 \pm 2.2$ \\
\textsc{Series-3} & $\mathbf{82.1} \pm 0.4$ & $85.6 \pm 0.2$ \\
\bottomrule
\end{tabular}
\caption{Test scores on ACE dataset with standard error. See Table~\ref{tbl:granular} for more details. \textsuperscript{*} is not fully comparable given the difference of encoder used.}\label{tbl:ace}
\vspace{-3mm}
\end{table}

\begin{table}[h]
\small
\centering
\begin{tabular}{l|l|l}
\toprule
Prompt & Micro-F1 & Macro-F1 \\
\midrule
\textsc{JointArg} & $79.9$ & - \\
\textsc{SpecialTokens} & $57.1 \pm 1.7$ & $55.8 \pm 2.5$ \\
\textsc{Name} & $\mathbf{82.0} \pm 0.1$ & $\mathbf{81.2} \pm 0.1$ \\
\textsc{Expert} & $81.2 \pm 0.1$ & $81.0 \pm 0.1$ \\
\textul{\textsc{Series-1}} & $\mathbf{82.0} \pm 0.0$ & $\mathbf{81.3} \pm 0.1$ \\
~~- \textsc{Pretraining} & $78.6 \pm 0.5$ & $77.7 \pm 0.3$ \\
\bottomrule
\end{tabular}
\caption{Test scores on RAMS dataset with standard error. See Table~\ref{tbl:granular} for further details.}\label{tbl:rams}
\vspace{-4mm}
\end{table}

\begin{figure*}[ht]
\begin{center}
    \includegraphics[width=.95\textwidth]{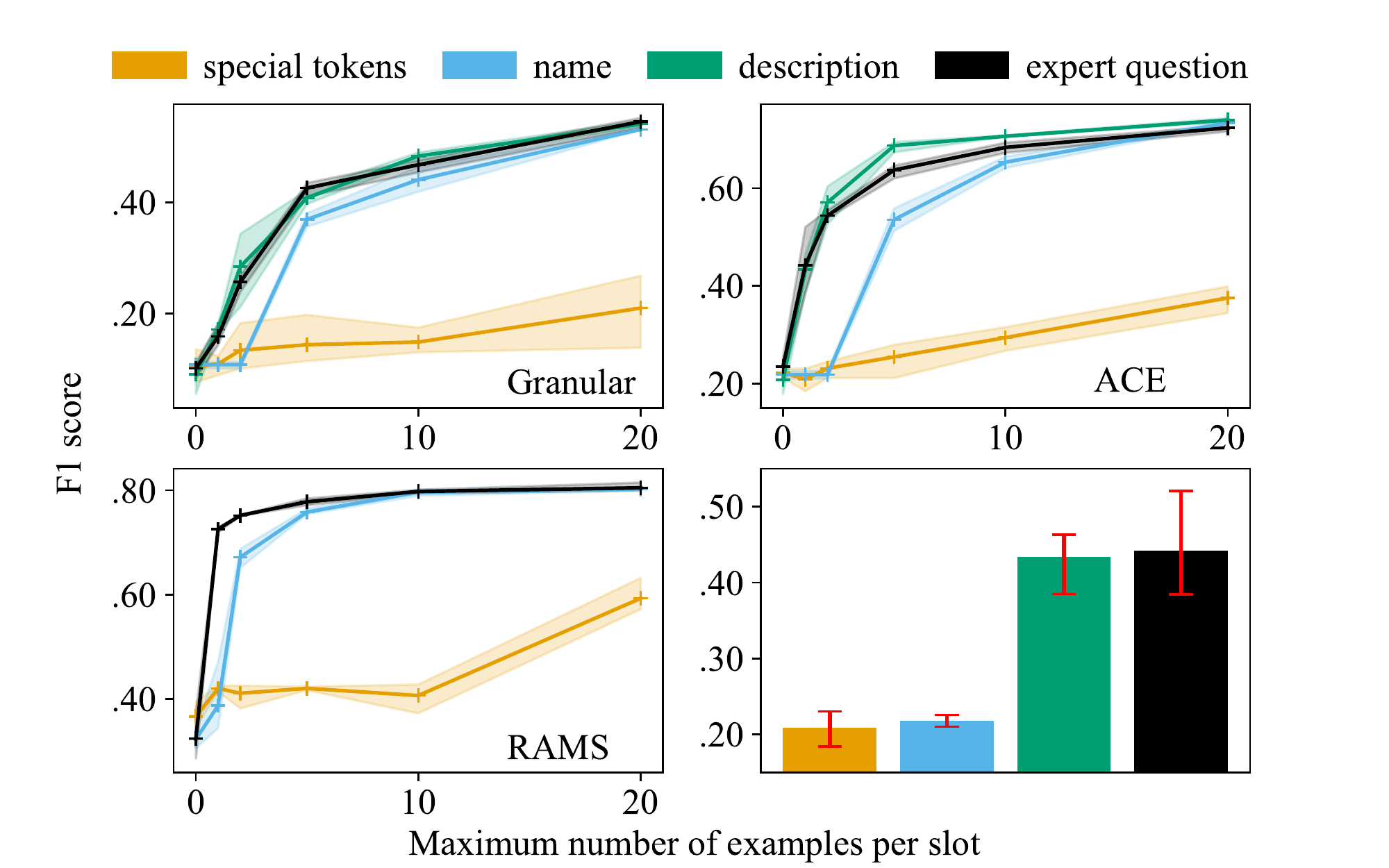}
\end{center}
\caption{Granular, ACE and RAMS Micro F1 as a function of slot frequency in the training data. Each data point is an average over 3 training runs with different random seeds. Bands and error bars indicate the full range of values. The bottom right bar plot corresponds to ACE, with at most 1 positive and 1 negative example per slot.}
\label{fig:graph-few-shot}
\vspace{-3mm}
\end{figure*}

Overall, we find that the model is able to leverage the information contained in NL prompts, as evidenced by the lower performance of \textsc{SpecialTokens}.
The best performance is consistently achieved by prompts formulated as questions, significantly better than the fluent \textsc{Description}.

While the expert question achieves good performance on each dataset, results suggest that knowledge of TE is not necessary to write questions that yield good performance.

Surprisingly, \textsc{Name} performs consistently well, being comparable to the best-performing question on 2 out of 3 datasets.
\textsc{Name} consists of the template name and the slot name, separated by one space.
Since slot names may be shared across templates, and since models can use the slot name as a proxy for the kind of answer expected, this could help the model spot the correct answer, simply based on surface patterns.
Consistent with our findings, \citet{dai19deeper} show, in the context of Information Retrieval, that whether a keyword-based or a fluent query performs best can depend on both the dataset the retrieval is performed on, and the retrieval method being used.

We further experiment with few-shot learning by subsampling positive and negative examples in the training data so that they appear less than a certain number of times (see Figure~\ref{fig:graph-few-shot}).
The observations above extend to the few-shot setting: the expert question outperforms other prompts for most data regimes. In addition, we find that as above, \textsc{Name} consistently achieves good performance, although with a slightly shallower learning curve.

Finally, the aforementioned Unified-QA encoder is built on T5-Base.
In Tables~\ref{tbl:granular} to~\ref{tbl:rams}, ``-~\textsc{Pretraining}'' reports the results of training the model on the line above, but using T5-Base as the encoder, thus depriving it of pretraining on QA datasets.
This ablation experiment allows us to examine the effect of QA pretraining, and shows that a large part of the performance indeed comes from QA-pretraining.
In fact, the ACE, RAMS and Granular datasets lie, in that order, along a spectrum stretching from sentence-level TE to document-level TE. The type of QA pretraining we have used here involves mostly sentence-level, factoid questions~\citep{khashabi20unifiedqa}. As expected, it helps more for sentence-level TE, and less for document-level TE.
While it is always possible to map TE to a QA problem, a QA-pretrained encoder is not necessarily the best-performing encoder.

The datasets used here rely on general-domain ontologies, so that the annotators possess common sense knowledge about their templates and slots. The conclusions of this study might be different if instead, the datasets relied on specialized ontologies, which the annotators would not be familiar with, such as ontologies in the legal or medical domain. Such experiments are out of scope here, and we leave it to future work.

As can be seen in Table~\ref{tbl:granular}, picking the right set of questions is crucial for final performance.
This is more challenging in the low-resource scenario, with little validation data.

\section{Analysis of human judgements}
\label{sec:scoring-questions}
Can humans predict performance on TE-QA?
Can humans reliably pick the best-performing question for a given slot, effectively replacing the role of the dev set?
For each slot, we asked human annotators to rank questions according to how well they describe the slot, without access to data or experimental results.

\mysubsection{Annotation protocol}

For each slot of each template in the Granular ontology, we presented human annotators with all questions written for that slot.\footnote{Annotators come from the same demographics as the annotators of Section~\ref{sec:writing-questions} and were compensated in the same way for their work.}
We asked them to rate each question on a scale of 1 to 4:

\begin{enumerate}[topsep=0pt,itemsep=0pt,partopsep=0pt,parsep=0pt]
    \item This is the best out of all the questions
    \item This is not the best, but among the best
    \item This is not the worst, but among the worst
    \item This is the worst out of all the questions
\end{enumerate}
Annotators were allowed to give the same rating to multiple questions, especially since some questions can be quite similar.
The annotators were shown the template description as well as the slot description.
There is no overlap between the annotators writing questions and those scoring questions.
We obtain 3 ratings per question for the Granular dataset.
We performed this study on the Granular dataset only, since performance achieved by different sets of questions was noticeably different.
Examples from the annotation process are shown in Table~\ref{tbl:scoring}.

\begin{table}[h!]
\small
\centering
\setlength{\tabcolsep}{0.01\linewidth}
\begin{tabular}{p{0.016\linewidth} p{0.016\linewidth}p{0.016\linewidth}|p{0.05\linewidth}|p{0.8\linewidth}}
\toprule
\multicolumn{5}{l}{Human Ratings (3) --- Aggregate Score --- Question} \\
\midrule
\multicolumn{5}{p{0.95\linewidth}}{\textsc{Epidemiplate}. The Epidemiplate template captures key information elements in regards to various kinds of disease outbreaks.} \\
\multicolumn{5}{p{0.95\linewidth}}{\textsc{Disease}. Mentions of the disease that is at the heart of the disease outbreak.} \\
\midrule
1 & 1 & 1 & 2.0 & What disease caused the outbreak? \\
2 & 1 & 1 & 1.5 & What is the disease causing the outbreak?  \\
3 & 2 & 3 & 1.0 & \emph{What disease broke out?}             \\
4 & 4 & 4 & 0.8 & What is the disease?                      \\
\midrule
\multicolumn{5}{p{0.95\linewidth}}{\textsc{Protestplate}. This Granular event is intended to capture the various aspects of a protest event.} \\
\multicolumn{5}{p{0.95\linewidth}}{\textsc{Arrested}. Description or count of those arrested.} \\
\midrule
1 & 1 & 2 & 2.2 & Who/How many were arrested at the protests?  \\
2 & 3 & 1 & 1.4 & Who was arrested because of the protest? \\
3 & 2 & 3 & 0.7 & Who and how many were arrested?  \\
3 & 2 & 3 & 0.7 & Who and/or how many people have been arrested? \\
2 & 3 & 2 & 0.7 & Who was arrested during or after the protest?  \\
2 & 3 & 2 & 0.7 & Who was arrested in the protest?\\
4 & 4 & 4 & 0.4 & \emph{Who was arrested?} \\
\bottomrule
\end{tabular}
\caption{Example question ratings from the Granular dataset. The expert question is in \emph{italics}.}\label{tbl:scoring}
\vspace{-4mm}
\end{table}

For a given slot, the \emph{ratings} of each annotator implicitly established a \emph{ranking} of the questions, which we use to compute an aggregated score for each question.
Using the \emph{rankings} instead of the \emph{ratings} normalizes the annotators' ratings.
We assign each question a score of $\sum_{i=1}^{3} \frac{1}{r_i}$ where $r_i$ is the rank attributed by the \mbox{$i$-th} annotator.
For questions with identical ranking, we picked the lowest possible rank --- e.g. 3 questions ranked first received each a score of~$\frac{1}{3}$.
Using these scores, we can pick a \textsc{Best} and \textsc{Worst} question for each slot, as the questions that get the highest and lowest score, respectively.
Examples of scores are shown in Table~\ref{tbl:scoring}.

The scores provided by the 3 annotators have an average Pearson correlation of 31.4\%.
The scores on questions implicitly provided a scoring of the annotators who wrote the questions.
We assigned each question-writer the average score of their questions.
The scores of question-writers ranged from 1.1 to 0.8, with the expert rated a 0.9, and ranked 5th out of 9.

Looking through the prompt set labeled \textsc{Best}, we find that annotators seem to have favored questions that are grammatically correct and fluent, but with unconventional typesetting.
For instance, 6 questions contain a ``/'', as in \emph{What did this person/network do that was corrupt?} and \emph{Who and/or how many people have been injured in the protest?}.
Similarly, 2 questions contain a ``(s)'' appended to the end of a word, as in \emph{What building(s) and space(s) are the protestors occupying?}.
In contrast, the expert questions contain neither of those, and the best performing set contains 4 ``/'' and no ``(s)''.
The encoder's tokenizer, built using a form of aggregation of characters and sub-word units, keeps certain frequent words whole, but breaks up less frequent ones.
Taking an example from above: \emph{What did this person do that was corrupt?} is tokenized as [\texttt{What}, \texttt{did}, \texttt{this}, \texttt{person}, \texttt{do}, \texttt{that}, \texttt{was}, \texttt{corrupt}, \texttt{?}] (and similarly if replacing ``person'' with ``network'').
In contrast, \emph{What did this person/network do that was corrupt?} is tokenized as [\texttt{What}, \texttt{did}, \texttt{this}, \texttt{person}, \texttt{/}, \texttt{net}, \texttt{work}, \texttt{do}, \texttt{that}, \texttt{was}, \texttt{corrupt}, \texttt{?}].
With conventional typesetting, the tokenizer keeps occurrences of the word ``network'' as a single token. The encoder has thus rarely, if at all, seen the token \texttt{net} followed by the token \texttt{work}, and likely cannot compose their meaning back into that of the token \texttt{network}.
More generally, every time the tokenizer is forced to back off to a sub-word representation, word-level information potentially useful for retrieval may be lost.

\mysubsection{Experiments}

As shown in Table~\ref{tbl:granular}, performance achieved by \textsc{Best} and \textsc{Worst} questions is within that of other question prompts, neither standing out as particularly good or particularly bad.
To quantify more precisely whether human scores could predict F1 score performance, we ran our models trained on questions on the test set, swapping in prompts they were not trained on.
We then computed, for each given model, the Pearson correlation between F1 performance (averaged over 5 runs), and the score attributed by humans to the question.
We found that the correlation was less than .1 in absolute value, so that overall, human preferences do not correlate with performance.

\section{Conclusion}
\label{sec:conclusion}
We show that NL questions written by potential end users of a TE system consistently achieve the best performance on TE-QA on all three datasets.
This finding extends to few-shot learning.
In practice, we find that a technical background in TE is not strictly necessary to formulate good questions, meaning that end users of the TE system can be trusted to drive the task definition by writing questions.
However, humans are unable to estimate the usefulness of a particular question phrasing with respect to TE-QA model performance.
Our results encourage the data-driven combination of different questions to achieve better performance on TE-QA.
In addition, as TE is shifting to a low resource scenario in which ontologies are built but not annotated for, our results suggest that designers of such ontologies ought to include, for each template and slot, questions that best describe the information meant to match the template and slot.

\section*{Acknowledgements}
We thank Tongfei Chen, Zhengping Jiang, Dawn Lawrie, Doug Oard, Guanghui Qin, Kate Sanders, Nathaniel Weir, and Patrick Xia for helpful comments and feedback.

\bibliography{anthology,custom}
\bibliographystyle{acl_natbib}

\appendix

\section{Appendices}
\label{sec:appendix}


\begin{table*}[h!]
\small
\centering
\begin{tabular}{p{2cm}|l|p{3cm}|l}
\toprule
part of model               & hyper-parameter name       & description & value range \\
\midrule
encoder              & model                     & Transformer model encoding the paragraph and prompt & \texttt{allenai/unifiedqa-t5-base} \\
                            & freeze                    & Whether to freeze the encoder's parameters & False \\
\midrule
MLP input features   & distance feature size     & size of the embeddings encoding the distance of each token to the template anchor & \{8, 16, 32, 64,128, 256, 512\} \\
                            & prior type                & the type of prior to use to compute the scores for each slot & \{none, embed, logit\}\\
\midrule
MLP classifier       & dropout                   & feature dropout in between the layers of the MLP & [0, 0.8] \\
                            & number of units           & - & \{32, 64, 128, 256, 512, 1024, 2048\} \\
                            & number of layers          & - & \{1, 2, 3, 4\} \\
\midrule
trainer              & batch size                & - & \{8, 16, 32, 64, 128, 256\} \\
                            & learning rate             & - & [1e-6, 1e-3] \\
                            & learning rate scheduler   & This specific scheduler halves the learning rate when the validtion score stops improving & \texttt{reduce\_on\_plateau} \\
                            & optimizer                 & - & \texttt{huggingface\_adamw} \\
                            & number of epochs          & Maximum number of epochs to train for & 25 \\
                            & patience                  & - & 5 \\
\midrule
loss function        & loss trade-off            & see $\lambda$ in Section~\ref{sec:model} & [0, 1]\\
                            & negative margin           & see $\beta$ in Section~\ref{sec:model} & [1e-2, 0.5] \\
                            & positive margin           & see $\alpha$ in Section~\ref{sec:model} & [1e-2, 0.5] \\
\bottomrule
\end{tabular}
\caption{Hyper-parameters used for the hyper-parameter search. We use Optuna~\citep{akiba19optuna} to find good hyper-parameters for each set of prompts, running 20 experiments per set. If there is a single value under ``value range'', it means this parameters was set and not searched over with Optuna. Each experiment consists of 3 training runs with 3 different random seeds, using the average micro F1 score on the validation set; except for RAMS, where we use single runs. We compute the test score of the best set of hyper-parameters by running 5 training runs with distinct random seeds and taking the average score.}
\label{tbl:hyperparameters}
\end{table*}

\begin{table*}[h!]
\small
\centering
\begin{tabular}{l|c|c|c|c|c|c}
\toprule
 & \multicolumn{3}{c|}{micro} & \multicolumn{3}{c}{macro} \\
prompt set & P & R & F1 & P & R & F1 \\
\midrule
\textsc{SpecialTokens} & $27.8 \pm 2.9$ & $49.9 \pm 3.2$ & $35.3 \pm 3.0$ & $38.6 \pm 2.4$ & $41.5 \pm 3.1$ & $35.5 \pm 2.8$ \\
\textsc{Name} & $\mathbf{68.3} \pm 1.3$ & $47.2 \pm 0.5$ & $55.8 \pm 0.4$ & $\mathbf{69.1} \pm 1.4$ & $53.4 \pm 1.1$ & $\mathbf{57.3} \pm 1.0$ \\
\textsc{Description} & $57.8 \pm 2.3$ & $64.5 \pm 2.1$ & $60.6 \pm 0.7$ & $58.4 \pm 2.5$ & $55.0 \pm 1.0$ & $53.0 \pm 1.6$ \\
\textsc{Expert} & $57.0 \pm 0.4$ & $\mathbf{66.8} \pm 0.6$ & $\mathbf{61.5} \pm 0.3$ & $59.9 \pm 1.2$ & $\mathbf{58.5} \pm 1.2$ & $55.6 \pm 1.1$ \\
\textsc{Series-1} & $54.9 \pm 2.5$ & $62.5 \pm 2.8$ & $57.9 \pm 0.7$ & $59.9 \pm 1.6$ & $58.0 \pm 1.4$ & $54.8 \pm 1.3$ \\
\textsc{Series-2} & $62.0 \pm 0.4$ & $60.6 \pm 1.1$ & $\mathbf{61.2} \pm 0.5$ & $64.7 \pm 1.0$ & $56.5 \pm 1.2$ & $56.6 \pm 0.9$ \\
\textsc{Series-3} & $50.5 \pm 0.8$ & $\mathbf{67.0} \pm 1.6$ & $57.5 \pm 0.5$ & $54.0 \pm 1.6$ & $54.1 \pm 0.9$ & $49.9 \pm 1.1$ \\
\textsc{Series-4} & $61.3 \pm 0.6$ & $60.7 \pm 0.5$ & $61.0 \pm 0.4$ & $\mathbf{67.2} \pm 1.9$ & $\mathbf{59.4} \pm 1.0$ & $\mathbf{59.5} \pm 1.2$ \\
\textsc{\textul{Series-5}} & $57.9 \pm 1.1$ & $\mathbf{67.2} \pm 1.8$ & $\mathbf{62.1} \pm 0.4$ & $62.5 \pm 1.6$ & $\mathbf{60.4} \pm 0.7$ & $\mathbf{57.9} \pm 0.9$ \\
\textsc{- Pretraining} & $59.1 \pm 0.6$ & $\mathbf{69.3} \pm 0.4$ & $\mathbf{63.8} \pm 0.3$ & $62.6 \pm 1.5$ & $\mathbf{60.5} \pm 1.0$ & $\mathbf{58.1} \pm 1.2$ \\
\textsc{Series-6} & $60.4 \pm 1.3$ & $63.1 \pm 1.1$ & $\mathbf{61.6} \pm 0.5$ & $62.2 \pm 0.9$ & $56.5 \pm 1.3$ & $56.3 \pm 0.9$ \\
\textsc{Best} & $61.7 \pm 1.1$ & $58.1 \pm 2.3$ & $59.6 \pm 0.7$ & $\mathbf{66.6} \pm 1.3$ & $57.2 \pm 1.7$ & $\mathbf{58.0} \pm 1.0$ \\
\textsc{Worst} & $56.0 \pm 2.1$ & $\mathbf{68.0} \pm 1.2$ & $\mathbf{61.2} \pm 1.0$ & $60.4 \pm 1.8$ & $\mathbf{59.8} \pm 0.8$ & $55.8 \pm 1.4$ \\
\bottomrule
\end{tabular}
\caption{Micro and macro averaged scores on Granular dataset with standard error. Best scores and scores that would be the best by changing 1 standard error are bolded. The question set with the highest dev micro F1 is \textul{underlined}.}
\label{tbl:granular-full}
\end{table*}

\begin{table*}[t]
\small
\centering
\begin{tabular}{l|c|c|c|c|c|c}
\toprule
 & \multicolumn{3}{c|}{micro} & \multicolumn{3}{c}{macro} \\
prompt set & P & R & F1 & P & R & F1 \\
\midrule \textsc{SpecialTokens} & $62.1 \pm 1.7$ & $68.7 \pm 1.6$ & $65.1 \pm 1.5$ & $72.0 \pm 0.6$ & $75.4 \pm 0.6$ & $72.6 \pm 0.6$ \\
\textsc{Name} & $75.7 \pm 0.4$ & $89.5 \pm 0.5$ & $\mathbf{82.0} \pm 0.3$ & $83.8 \pm 0.2$ & $\mathbf{91.0} \pm 0.2$ & $\mathbf{86.3} \pm 0.2$ \\
\textsc{Description} & $73.7 \pm 0.5$ & $89.3 \pm 0.5$ & $80.8 \pm 0.2$ & $83.3 \pm 0.3$ & $\mathbf{91.1} \pm 0.4$ & $85.9 \pm 0.2$ \\
\textsc{Expert} & $\mathbf{80.3} \pm 0.5$ & $84.0 \pm 0.7$ & $\mathbf{82.1} \pm 0.2$ & $\mathbf{86.1} \pm 0.3$ & $88.1 \pm 0.4$ & $\mathbf{86.2} \pm 0.2$ \\
\textsc{Series-1} & $78.5 \pm 1.0$ & $85.9 \pm 0.9$ & $\mathbf{82.0} \pm 0.4$ & $\mathbf{85.7} \pm 0.6$ & $89.2 \pm 0.7$ & $\mathbf{86.4} \pm 0.5$ \\
\textul{\textsc{Series-2}} & $\mathbf{79.9} \pm 0.5$ & $84.9 \pm 0.9$ & $\mathbf{82.3} \pm 0.3$ & $\mathbf{86.5} \pm 0.5$ & $88.9 \pm 0.6$ & $\mathbf{86.7} \pm 0.4$ \\
\textsc{- Pretraining} & $77.7 \pm 1.7$ & $38.8 \pm 2.9$ & $51.6 \pm 2.9$ & $77.9 \pm 1.9$ & $67.0 \pm 2.1$ & $68.8 \pm 2.2$ \\
\textsc{Series-3} & $75.2 \pm 0.8$ & $\mathbf{90.5} \pm 0.3$ & $\mathbf{82.1} \pm 0.4$ & $83.3 \pm 0.2$ & $90.1 \pm 0.3$ & $85.6 \pm 0.2$ \\
\bottomrule
\end{tabular}
\caption{Micro and macro averaged scores on ACE dataset with standard error. See Table~\ref{tbl:granular-full} for further details.}\label{tbl:ace-full}
\end{table*}

\begin{table*}[h!]
\small
\centering
\begin{tabular}{l|c|c|c|c|c|c}
\toprule
 & \multicolumn{3}{c|}{micro} & \multicolumn{3}{c}{macro} \\
prompt set & P & R & F1 & P & R & F1 \\
\midrule \textsc{SpecialTokens} & $45.8 \pm 2.4$ & $76.6 \pm 1.1$ & $57.1 \pm 1.7$ & $50.2 \pm 3.1$ & $72.7 \pm 0.9$ & $55.8 \pm 2.5$ \\
\textsc{Name} & $\mathbf{81.5} \pm 0.6$ & $82.5 \pm 0.7$ & $\mathbf{82.0} \pm 0.1$ & $\mathbf{82.8} \pm 0.6$ & $82.4 \pm 0.3$ & $\mathbf{81.2} \pm 0.1$ \\
\textsc{Expert} & $73.5 \pm 0.2$ & $\mathbf{90.8} \pm 0.2$ & $81.2 \pm 0.1$ & $77.8 \pm 0.2$ & $\mathbf{88.2} \pm 0.1$ & $81.0 \pm 0.1$ \\
\textul{\textsc{Series-1}} & $75.1 \pm 0.2$ & $90.3 \pm 0.2$ & $\mathbf{82.0} \pm 0.0$ & $78.9 \pm 0.2$ & $87.5 \pm 0.2$ & $\mathbf{81.3} \pm 0.1$ \\
\textsc{- Pretraining} & $69.2 \pm 0.8$ & $\mathbf{91.0} \pm 0.6$ & $78.6 \pm 0.5$ & $73.2 \pm 0.6$ & $87.6 \pm 0.5$ & $77.7 \pm 0.3$ \\
\bottomrule
\end{tabular}
\caption{Average micro and macro scores for the RAMS dataset, with standard error of the mean. Statistics are computed over 5 runs of the best set of hyperparameters, each with a different random seed. Best scores are bolded. Scores that would be best scores if removing or adding one standard error of the mean are bolded as well. The prompt set that would have been picked based on the dev micro F1 score is in italics.}\label{tbl:rams-full}
\end{table*}

\begin{figure*}[h!]
\centering
\includegraphics[width=\textwidth]{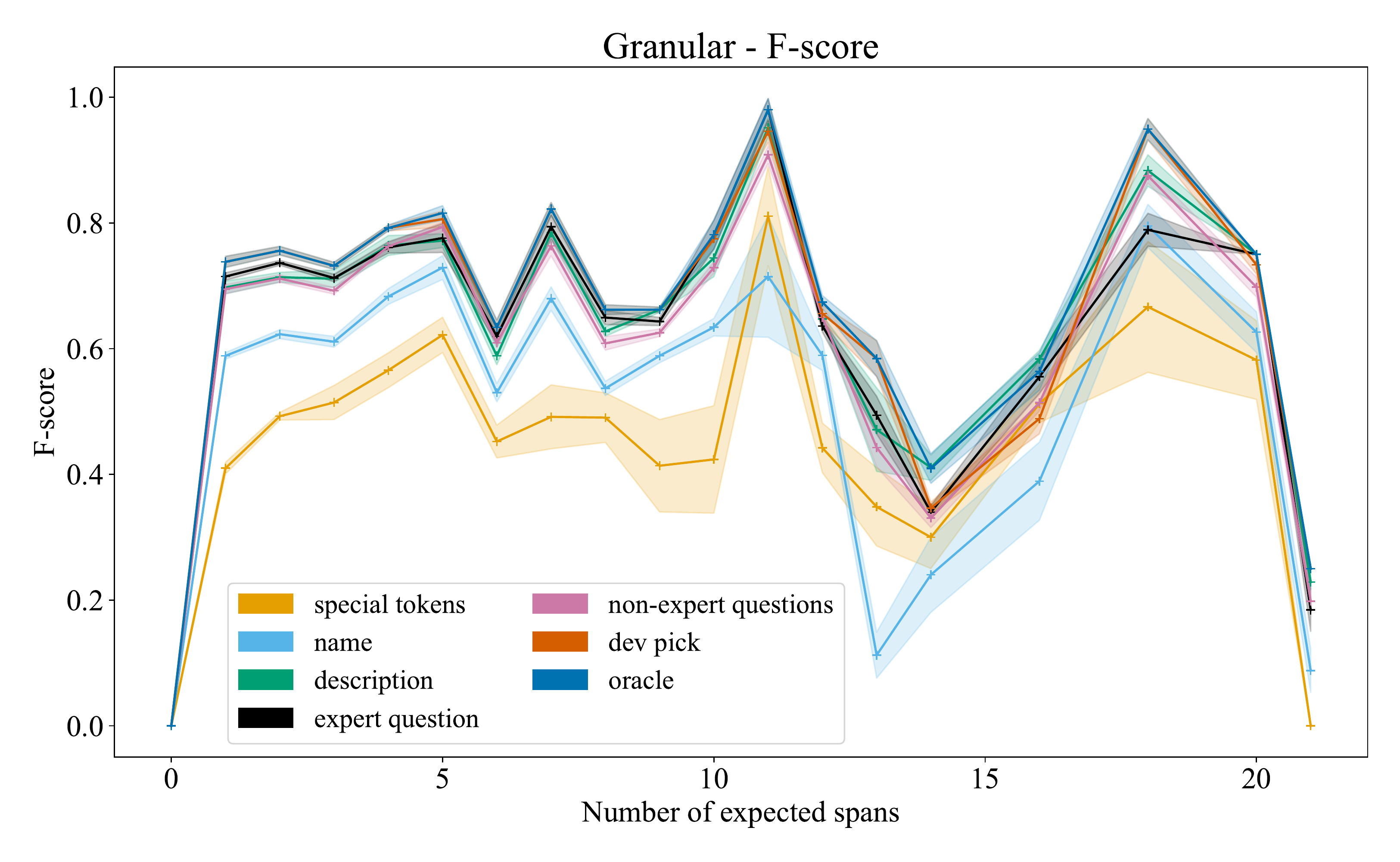}
\caption{F-score as a function of the number spans in the gold answer on the Granular dataset. Each point is an average over 5 runs. Bands indicate one standard error of the mean around the average.}
\label{fig:granular-line}
\end{figure*}

\begin{figure*}[h!]
\centering
\includegraphics[width=\textwidth]{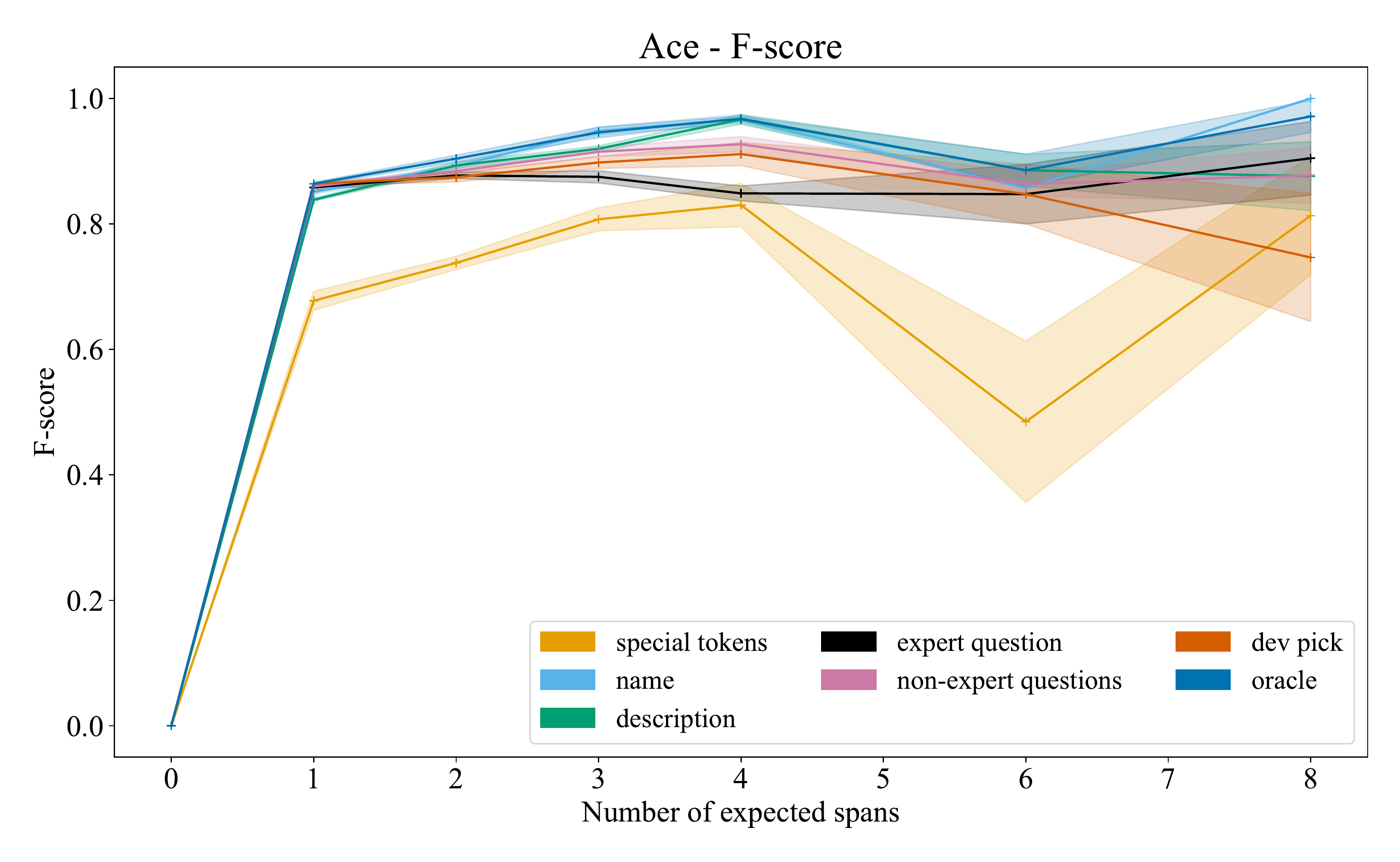}
\caption{F-score as a function of the number spans in the gold answer on the ACE dataset. Each point is an average over 5 runs. Bands indicate one standard error of the mean around the average.}
\label{fig:ace-line}
\end{figure*}

\begin{figure*}[h!]
\centering
\includegraphics[width=\textwidth]{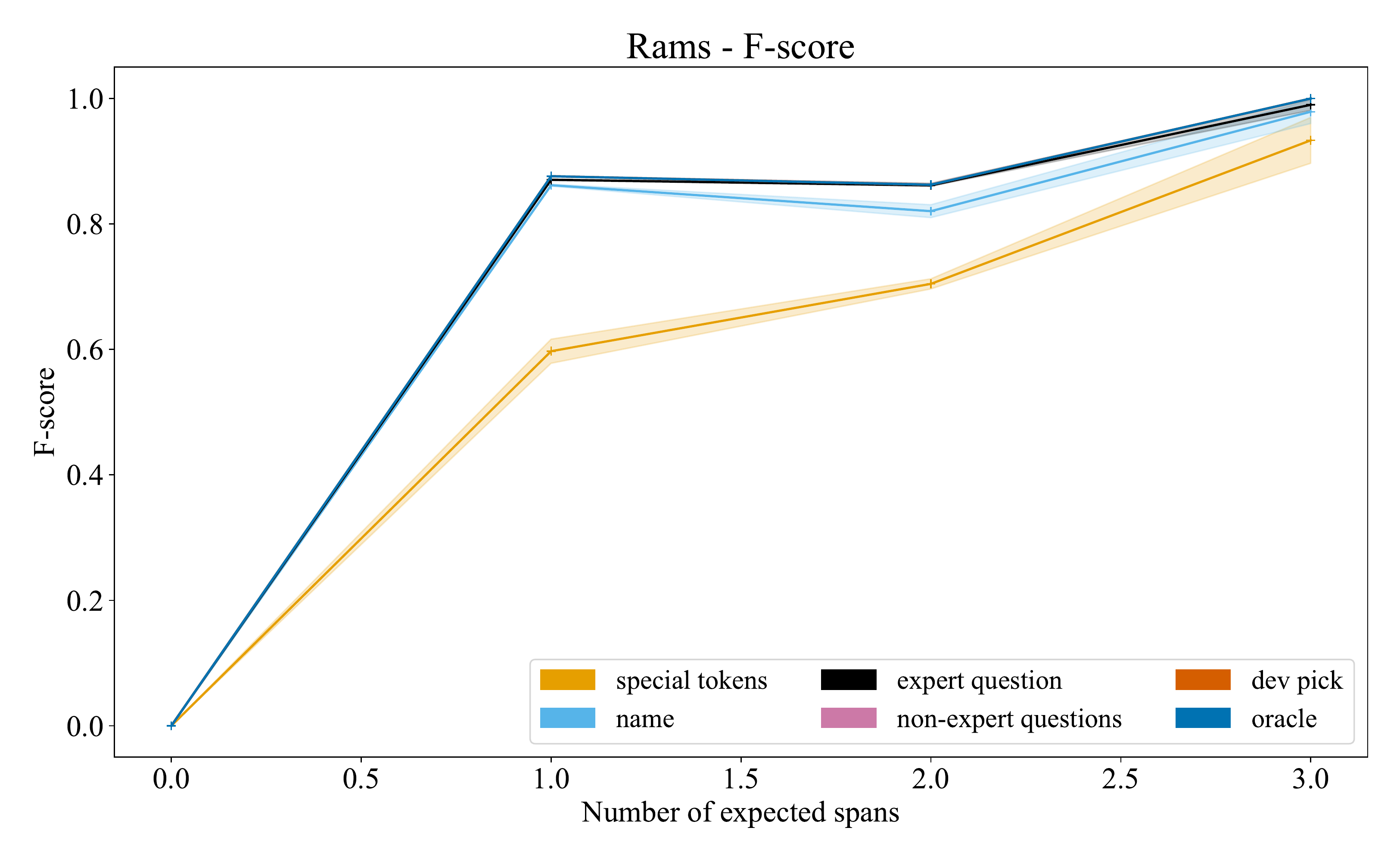}
\caption{F-score as a function of the number spans in the gold answer on the RAMS dataset. Each point is an average over 5 runs. Bands indicate one standard error of the mean around the average.}
\label{fig:rams-line}
\end{figure*}

\end{document}